\documentclass[10pt,twocolumn,letterpaper]{article}

\usepackage{cvpr}
\usepackage{times}
\usepackage{epsfig}
\usepackage{graphicx}
\usepackage{amsmath}
\usepackage{amssymb}
\usepackage{xcolor}
\usepackage{gensymb}

\usepackage{url}            
\usepackage{booktabs}       
\usepackage{amsfonts}       
\usepackage{nicefrac}       
\usepackage{microtype}      

\usepackage{times}
\usepackage{epsfig}
\usepackage{graphicx}
\usepackage{amsmath}
\usepackage{amssymb}

\usepackage{units}
\usepackage{multirow}
\usepackage{bm}
\usepackage{bbm}
\usepackage{pifont}
\usepackage{xspace} 

\usepackage{graphicx,calc} 
\usepackage{color}    
\usepackage{ifthen}
\usepackage{paralist}
\usepackage{times}
\usepackage{longtable}
\usepackage{colortbl}
\usepackage{nomencl}
\usepackage{bm}

\usepackage{amsmath}

\usepackage{authblk}

\DeclareMathOperator{\atantwo}{atan2}

\usepackage[pagebackref=true,breaklinks=true,letterpaper=true,colorlinks,bookmarks=false]{hyperref}

\def\cvprPaperID{****} 

\newcommand{\todo}[1]{}
\renewcommand{\todo}[1]{{\color{red} TODO: {#1}}}

\ifcvprfinal\fi

\title{Scalability in Perception for Autonomous Driving: Waymo Open Dataset}

\author[1]{Pei Sun}                                                                                                
\author[1]{Henrik Kretzschmar}                                                                                                               
\author[1]{Xerxes Dotiwalla}                                                                                                                 
\author[1]{Aur\'elien Chouard}                                                                                                                 
\author[1]{Vijaysai Patnaik}                                                                                                                 
\author[1]{Paul Tsui}                                                                                                                        
\author[1]{James Guo}                                                                                                                        
\author[1]{Yin Zhou}                                                                                                                         
\author[1]{Yuning Chai}                                                                                                                      
\author[2]{Benjamin Caine}                                                                                                                        
\author[2]{Vijay Vasudevan}                                                                                                                  
\author[2]{Wei Han}                                                                                                                          
\author[2]{Jiquan Ngiam}                                                                                                                     
\author[1]{Hang Zhao}                                                                                                                        
\author[1]{Aleksei Timofeev}                                                                                                                 
\author[1]{Scott Ettinger}                                                                                                                   
\author[1]{Maxim Krivokon}                                                                                                                   
\author[1]{Amy Gao}                                                                                                                          
\author[1]{Aditya Joshi}     

\author[1]{Sheng Zhao}

\author[1]{Shuyang Cheng}
\author[1]{Yu Zhang\thanks{Work done while at Waymo LLC.}}
\author[2]{Jonathon Shlens}                                                                                                                       
\author[2]{Zhifeng Chen}                                                                                                                     
\author[1]{Dragomir Anguelov}    
\makeatletter 
\renewcommand\AB@affilsepx{\quad \protect\Affilfont} 
\makeatother
\affil[1]{Waymo LLC}
\affil[2]{Google LLC}

\begin{document}
\maketitle

\begin{abstract}
The research community has increasing interest in autonomous driving research, despite the resource intensity of obtaining representative real world data. Existing self-driving datasets are limited in the scale and variation of the environments they capture, even though generalization within and between operating regions is crucial to the overall viability of the technology. In an effort to help align the research community's contributions with real-world self-driving problems, we introduce a new large-scale, high quality, diverse dataset. Our new dataset consists of 1150 scenes that each span 20 seconds, consisting of well synchronized and calibrated high quality LiDAR and camera data captured across a range of urban and suburban geographies. It is 15x more diverse than the largest camera+LiDAR dataset available based on our proposed geographical coverage metric. We exhaustively annotated this data with 2D (camera image) and 3D (LiDAR) bounding boxes, with consistent identifiers across frames. Finally, we provide strong baselines for 2D as well as 3D detection and tracking tasks. We further study the effects of dataset size and generalization across geographies on 3D detection methods. Find data, code and more up-to-date information at \ifcvprfinal{http://www.waymo.com/open}\else{http://anonymized}\fi.

\end{abstract}

\section{Introduction}
\label{sec:intro}

Autonomous driving technology is expected to enable a wide range of applications that have the potential to save many human lives, ranging from robotaxis to self-driving trucks.
The availability of public large-scale datasets and benchmarks has greatly accelerated progress in machine perception tasks, including image classification, object detection, object tracking, semantic segmentation as well as instance segmentation \cite{deng2009imagenet,lin2014microsoft,zhou2017scene,gupta2019lvis}.

To further accelerate the development of autonomous driving technology, we present the largest and most diverse multimodal autonomous driving dataset to date, comprising of images recorded by multiple high-resolution cameras and sensor readings from multiple high-quality LiDAR scanners mounted on a fleet of self-driving vehicles. The geographical area captured by our dataset is substantially larger than the area covered by any other comparable autonomous driving dataset, both in terms of absolute area coverage, and in distribution of that coverage across geographies. Data was recorded across a range of conditions in multiple cities, namely San Francisco, Phoenix, and Mountain View, with large geographic coverage within each city. We demonstrate that the differences in these geographies lead to a pronounced domain gap, enabling exciting research opportunities in the field of domain adaptation.


Our proposed dataset contains a large number of high-quality, manually annotated 3D ground truth bounding boxes for the LiDAR data, and 2D~tightly fitting bounding boxes for the camera images. All ground truth boxes contain track identifiers to support object tracking. In addition, researchers can extract 2D~amodal camera boxes from the 3D~LiDAR boxes using our provided rolling shutter aware projection library. The multimodal ground truth facilitates research in sensor fusion that leverages both the LiDAR and the camera annotations. Our dataset contains around 12~million LiDAR box annotations and around 12~million camera box annotations, giving rise to around 113k LiDAR object tracks and around 250k camera image tracks. All annotations were created and subsequently reviewed by trained labelers using production-level labeling tools.

We recorded all the sensor data of our dataset using an industrial-strength sensor suite consisting of multiple high-resolution cameras and multiple high-quality LiDAR sensors. Furthermore, we offer synchronization between the camera and the LiDAR readings, which offers interesting opportunities for cross-domain learning and transfer.  We release our LiDAR sensor readings in the form of range images. In addition to sensor features such as elongation, we provide each range image pixel with an accurate vehicle pose. This is the first dataset with such low-level, synchronized information available, making it easier to conduct research on LiDAR input representations other than the popular 3D point set format.

Our dataset currently consists of 1000~scenes for training and validation, and 150~scenes for testing, where each scene spans 20\,s. Selecting the test set scenes from a geographical holdout area allows us to evaluate how well models that were trained on our dataset generalize to previously unseen areas.

We present benchmark results of several state-of-the-art 2D-and 3D object detection and tracking methods on the dataset.

\section{Related Work}
\label{sec:related}

High-quality, large-scale datasets are crucial for autonomous driving research. There have been an increasing number of efforts in releasing datasets to the community in recent years.



Most autonomous driving systems fuse sensor readings from multiple sensors, including cameras, LiDAR, radar, GPS, wheel odometry, and IMUs. Recently released autonomous driving datasets have included sensor readings obtained by multiple sensors. Geiger \etal introduced the multi-sensor KITTI Dataset~\cite{REF:Geiger2012CVPR,REF:Geiger2013IJRR} in 2012, which provides synchronized stereo camera as well as LiDAR sensor data for 22~sequences, enabling tasks such as 3D object detection and tracking, visual odometry, and scene flow estimation. The SemanticKITTI Dataset~\cite{behley2019semantickitti} provides annotations that associate each LiDAR point with one of 28~semantic classes in all 22~sequences of the KITTI Dataset.

The ApolloScape Dataset~\cite{huang2018apolloscape}, released in 2017, provides per-pixel semantic annotations for 140k~camera images captured in various traffic conditions, ranging from simple scenes to more challenging scenes with many objects. The dataset further provides pose information with respect to static background point clouds.
The KAIST Multi-Spectral Dataset~\cite{choi2018kaist} groups scenes recorded by multiple sensors, including a thermal imaging camera, by time slot, such as daytime, nighttime, dusk, and dawn.
The Honda Research Institute 3D Dataset (H3D)~\cite{patil2019h3d} is a 3D~object detection and tracking dataset that provides 3D LiDAR sensor readings recorded in 160~crowded urban scenes.

Some recently published datasets also include map information about the environment. For instance, in addition to multiple sensors such as cameras, LiDAR, and radar, the nuScenes Dataset~\cite{REF:nuscenes2019} provides rasterized top-down semantic maps of the relevant areas that encode information about driveable areas and sidewalks for 1k~scenes. This dataset has limited LiDAR sensor quality with ~34K points per frame, limited geographical diversity covering an effective area of $5\textrm{km}^2$ (Table ~\ref{Dataset_comparison}).

In addition to rasterized maps, the Argoverse Dataset~\cite{REF:argoverse_CVPR2019} contributes detailed geometric and semantic maps of the environment comprising information about the ground height together with a vector representation of road lanes and their connectivity. They further study the influence of the provided map context on autonomous driving tasks, including 3D tracking and trajectory prediction. Argoverse has a very limited amount raw sensor data released.

See Table~\ref{Dataset_comparison} for a comparison of different datasets.

\begin{table}[!tbp]
\footnotesize
\begin{center}
\begin{tabular}{lcccc}
\toprule
& KITTI &  NuScenes & Argo & Ours \\
\midrule
Scenes & 22 & 1000 & 113 & 1150 \\ 
Ann. Lidar Fr. & 15K & 40K & 22K & 230K \\
Hours &1.5 & 5.5 & 1 & 6.4 \\
\midrule
3D Boxes &80K & 1.4M & 993k & 12M \\ 
2D Boxes &80K & -- & -- & 9.9M \\
\midrule
Lidars & 1& 1 & 2 & 5 \\
Cameras & 4& 6 & 9 &5 \\
Avg Points/Frame & 120K & 34K & 107K & 177K \\
LiDAR Features & 1 & 1 & 1 & 2\\
\midrule

Maps & No & Yes & Yes & No \\
Visited Area $(\textrm{km}^2)$ &-- & 5 & 1.6 & 76 \\
\bottomrule

\end{tabular}
\end{center}
\caption{Comparison of some popular datasets. The Argo Dataset refers to their Tracking dataset only, not the Motion Forecasting dataset. 3D~labels projected to 2D are not counted in the 2D~Boxes. Avg Points/Frame is the number of points from all LiDAR returns computed on the released data. Visited area is measured by diluting trajectories by 75 meters in radius and union all the diluted areas. Key observations: 1. Our dataset has 15.2x effective geographical coverage defined by the diversity area metric in Section ~\ref{dataset_analysis}. 2. Our dataset is larger than other camera+LiDAR datasets by different metrics. (Section ~\ref{sec:related}) \vspace{0cm}}
\label{Dataset_comparison}
\end{table}

\begin{table}[!tbp]
\begin{center}
\begin{tabular}{lcc}
\toprule
& TOP & F,SL,SR,R \\
\midrule
VFOV &  [-17.6\degree, +2.4\degree] & [-90\degree, 30\degree] \\
Range (restricted) & 75 meters & 20 meters \\
Returns/shot & 2 & 2 \\
\bottomrule
\end{tabular}
\end{center}
\caption{LiDAR Data Specifications for Front (F), Right (R), Side-Left (SL), Side-Right (SR), and Top (TOP) sensors. The vertical field of view (VFOV) is specified based on inclination (Section ~\ref{sec:coorid_system}). \vspace{-0.4cm}}
\label{LiDAR_Specs}
\end{table}

\begin{table}[!tbp]
\begin{center}
\begin{tabular}{lccc}
\toprule
& F & FL,FR & SL,SR \\
\midrule
Size &  1920x1280 & 1920x1280 & 1920x1040 \\
HFOV & $\pm25.2\degree$ & $\pm25.2\degree$ & $\pm25.2\degree$ \\
\bottomrule
\end{tabular}
\end{center}
\caption{\small{Camera Specifications for Front (F), Front-Left (FL), Front-Right (FR), Side-Left (SL), Side-Right (SR) cameras. The image sizes reflect the results of both cropping and downsampling the original sensor data.
The camera horizontal field of view (HFOV) is provided as an angle range in the x-axis in the x-y plane of camera sensor frame (Figure~\ref{fig:sensor_layout}). \vspace{-0.1cm}}}
\label{Camera_Specs}
\end{table}


\section{\ifcvprfinal{Waymo }\else{}\fi Open Dataset}
\label{sec:waymo_od}
\subsection{Sensor Specifications}
\begin{figure}[t!]
    \centering
    \includegraphics[width=1.1\columnwidth]{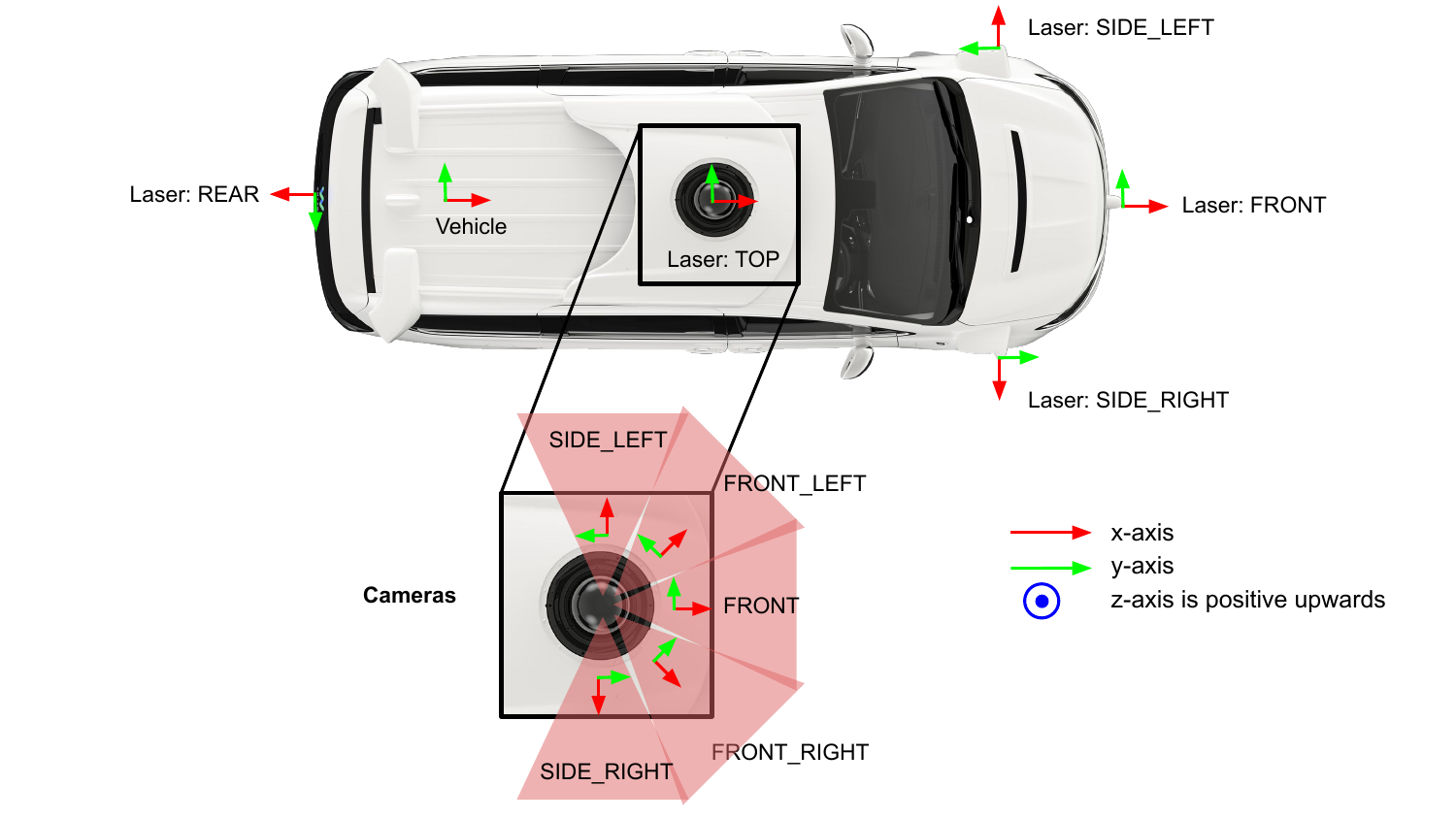}
    \caption{Sensor layout and coordinate systems.  \vspace{-0.4cm}}
    \label{fig:sensor_layout}
\end{figure}

The data collection was conducted using five LiDAR sensors and five high-resolution pinhole cameras. We restrict the range of the LiDAR data, and provide data for the first two returns of each laser pulse. Table~\ref{LiDAR_Specs} contains detailed specifications of our LiDAR data. The camera images are captured with rolling shutter scanning, where the exact scanning mode can vary from scene to scene. All camera images are downsampled and cropped from the raw images; Table~\ref{Camera_Specs} provides specifications of the camera images. See Figure ~\ref{fig:sensor_layout} for the layout of sensors relevant to the dataset.

\subsection{Coordinate Systems}
\label{sec:coorid_system}
This section describes the coordinate systems used in the dataset. All of the coordinate systems follow the right hand rule, and the dataset contains all information needed to transform data between any two frames within a run segment.

The \textbf{Global frame} is set prior to vehicle motion. It is an East-North-Up coordinate system: Up (z) is aligned with the gravity vector, positive upwards; East (x) points directly east along the line of latitude; North (y) points towards the north pole.

The \textbf{Vehicle frame} moves with the vehicle. Its x-axis is positive forwards, y-axis is positive to the left, z-axis is positive upwards. A vehicle pose is defined as a 4x4 transform matrix from the vehicle frame to the global frame. Global frame can be used as the proxy to transform between different vehicle frames. Transform among close frames is very accurate in this dataset.

A \textbf{Sensor frame} is defined for each sensor. It is denoted as a 4x4 transformation matrix that maps data from sensor frame to vehicle frame. This is also known as the "extrinsics" matrix.

The LiDAR sensor frame has z pointing upward. The x-y axes depends on the LiDAR.

The camera sensor frame is placed at the center of the lens. The x axis points down the lens barrel out of the lens. The z axis points up. The y/z plane is parallel to the image plane.

The \textbf{Image frame} is a 2D coordinate system defined for each camera image, where +x is along the image width (i.e. column index starting from the left), and +y is along the image height (i.e. row index starting from the top). The origin is the top-left corner.

The \textbf{LiDAR Spherical coordinate system} is based on the Cartesian coordinate system in the LiDAR sensor frame. A point (x, y, z) in the LiDAR Cartesian coordinate system can be uniquely transformed to a (range, azimuth, inclination) tuple in the LiDAR Spherical coordinate system by the following equations:
\begin{align}
    \textrm{range} & = \sqrt{x^2 + y^2 + z^2} \\
    \textrm{azimuth} & = \atantwo(y, x) \\
    \textrm{inclination} & = \atantwo(z, \sqrt{x^2 + y^2}).
\end{align}

\subsection{Ground Truth Labels}
\label{ground_truth_labels}
We provide high-quality ground truth annotations, both for the LiDAR sensor readings as well as the camera images. Separate annotations in LiDAR and camera data opens up exciting research avenues in sensor fusion. For any label, we define length, width, height to be the sizes along x-axis, y-axis and z-axis respectively.

\begin{figure}[t!]
    \centering
    \includegraphics[width=1\columnwidth]{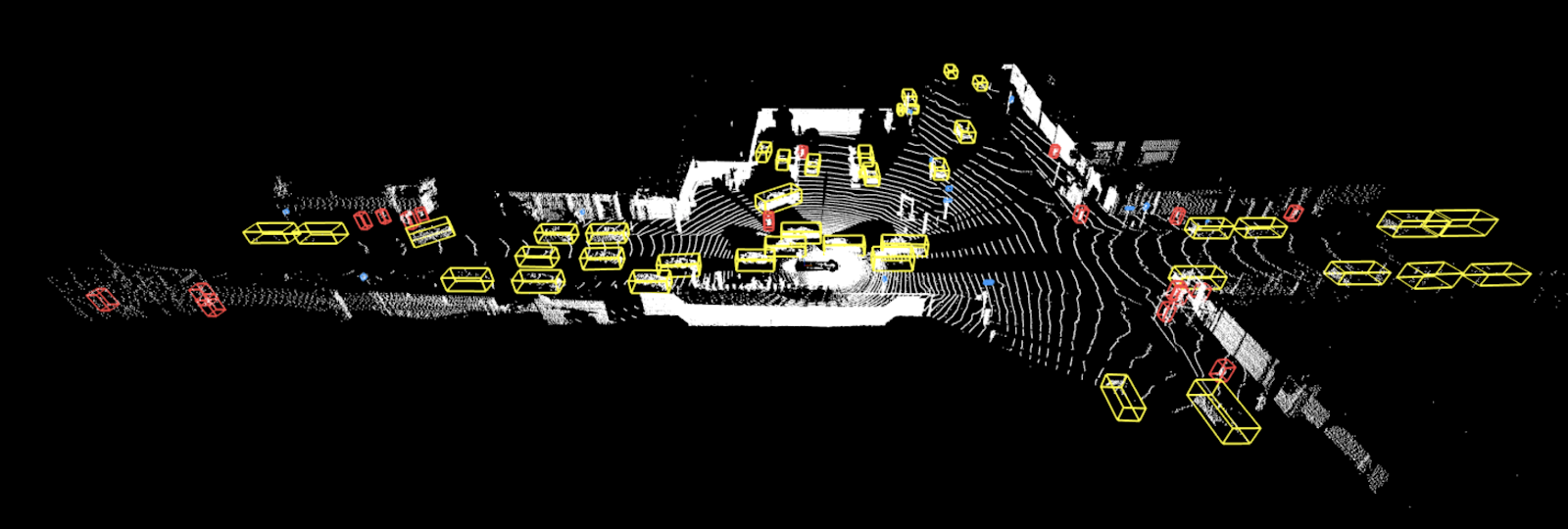}
    \caption{LiDAR label example. Yellow = vehicle. Red = pedestrian. Blue = sign. Pink = cyclist. \vspace{0cm}}
    \label{fig:LiDAR_label_ex}
\end{figure}

We exhaustively annotated vehicles, pedestrians, signs and cyclists in the LiDAR sensor readings. We labeled each object as a 7-DOF 3D upright bounding box (cx, cy, cz, l, w, h, $\theta$) with a unique tracking ID, where $cx$, $cy$, $cz$ represent the center coordinates, $l$, $w$, $h$ are the length, width, height, and $\alpha$ denotes the heading angle in radians of the bounding box. Figure~\ref{fig:LiDAR_label_ex} illustrates an annotated scene as an example.

In addition to the LiDAR labels, we separately exhaustively annotated vehicles, pedestrians and cyclists in all camera images. We annotated each object with a tightly fitting 4-DOF image axis-aligned 2D~bounding box which is complementary to the 3D boxes and their amodal 2D projections. The label is encoded as ($cx$, $cy$, $l$, $w$) with a unique tracking ID, where $cx$ and $cy$ represent the center pixel of the box, $l$ represents the length of the box along the horizontal (x) axis in the image frame, and $w$ represent the width of the box along the vertical (y) axis in the image frame. We use this convention for length and width to be consistent with 3D boxes. One interesting possibility that can be explored using the dataset is the prediction of 3D boxes using camera only.

We use two levels for difficulty ratings, similar to KITTI, where the metrics for LEVEL\_2 are cumulative and thus include LEVEL\_1. The criteria for an example to be in a specific difficulty level can depend on both the human labelers and the object statistics.



We emphasize that all LiDAR and all camera groundtruth labels were manually created by highly experienced human annotators using industrial-strength labeling tools. We have performed multiple phases of label verification to ensure a high labeling quality.


\subsection{Sensor Data}
\textbf{LiDAR data} is encoded in this dataset as range images, one for each LiDAR return; data for the first two returns is provided. The range image format is similar to the rolling shutter camera image in that it is filled in column-by-column from left to right. Each range image pixel corresponds to a LiDAR return. The height and width are determined by the resolution of the inclination and azimuth in the LiDAR sensor frame. Each inclination for each range image row is provided. Row 0 (the top row of the image) corresponds to the maximum inclination. Column 0 (left most column of the image) corresponds to the negative x-axis (\textit{i.e.,} the backward direction). The center of the image corresponds to the positive x-axis (\textit{i.e.,} the forward direction). An azimuth correction is needed to make sure the center of the range image corresponds to the positive x-axis.

\begin{figure}[t!]
    \centering
    \includegraphics[height=0.25\textwidth,width=1\columnwidth]{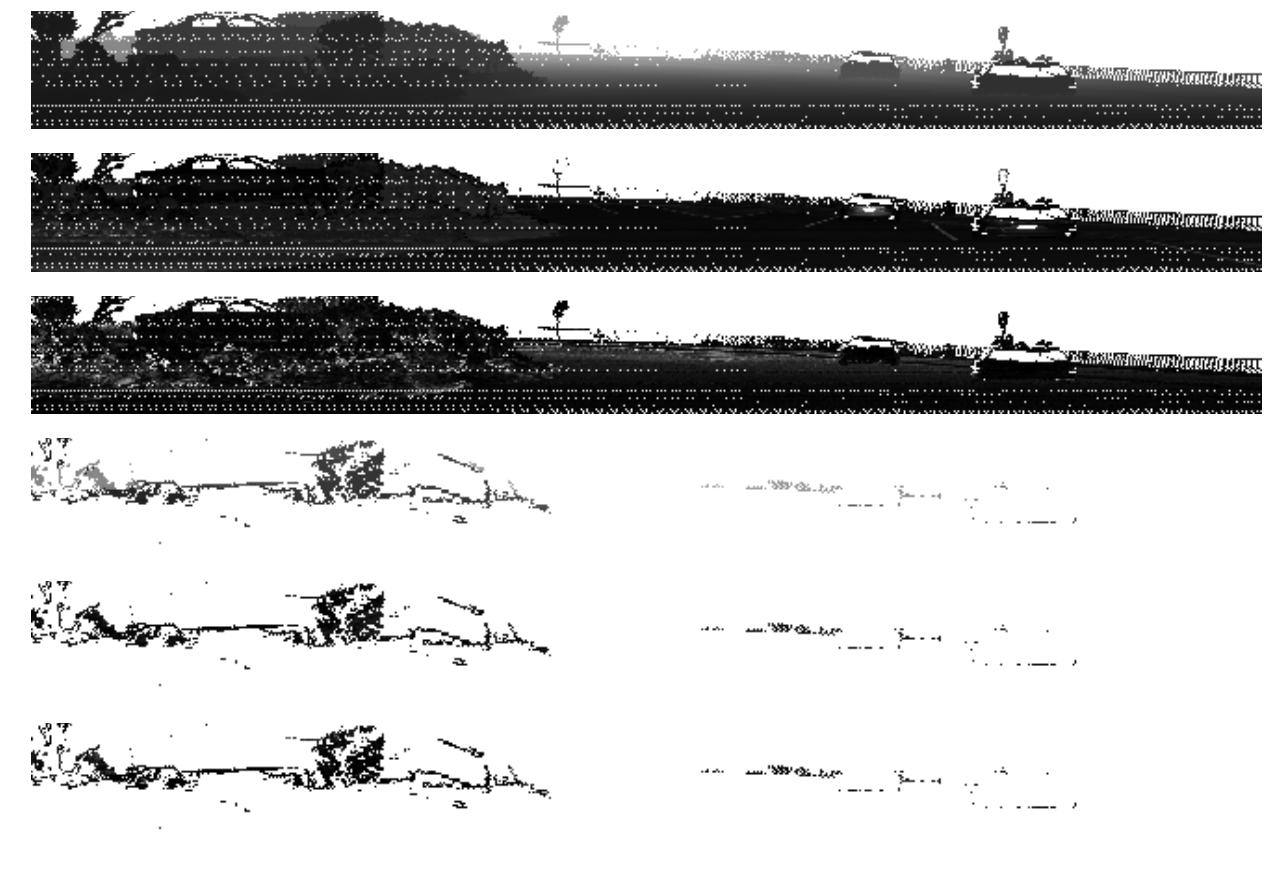}
    \caption{A range image example. It is cropped to only show the front 90\degree. The first three rows are range, intensity, and elongation from the first LiDAR return. The last three are range, intensity, and elongation from the second LiDAR return.\vspace{0cm}}
    \label{fig:ri_ex}
\end{figure}

\begin{figure}[t!]
    \centering
    \includegraphics[height=0.25\textwidth,width=1\columnwidth]{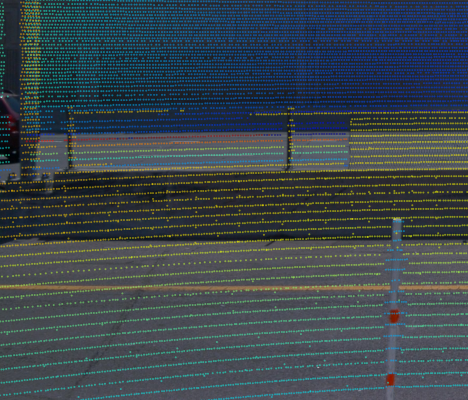}
    \caption{An example image overlaid with LiDAR point projections.\vspace{-0.4cm}}
    \label{fig:cp_ex}
\end{figure}

Each pixel in the range image includes the following properties. Figure \ref{fig:ri_ex} demonstrates an example range image.
\begin{compactitem}
\item Range: The distance between the LiDAR point and the origin in LiDAR sensor frame.
\item Intensity: A measurement indicating the return strength of the laser pulse that generated the LiDAR point, partly based on the reflectivity of the object struck by the laser pulse.
\item Elongation: The elongation of the laser pulse beyond its nominal width. Elongation in conjunction with intensity is useful for classifying spurious objects, such as dust, fog, rain. Our experiments suggest that a highly elongated low-intensity return is a strong indicator for a spurious object, while low intensity alone is not a sufficient signal.
\item No label zone: This field indicates whether the LiDAR point falls into a no label zone, \textit{i.e.,} an area that is ignored for labeling.
\item Vehicle pose: The pose at the time the LiDAR point is captured.
\item Camera projection: We provide accurate LiDAR point to camera image projections with rolling shutter effect compensated. Figure \ref{fig:cp_ex} demonstrates that LiDAR points can be accurately mapped to image pixels via the projections.
\end{compactitem}

\begin{figure}[t!]
    \centering
    \includegraphics[height=0.25\textwidth,width=1\columnwidth]{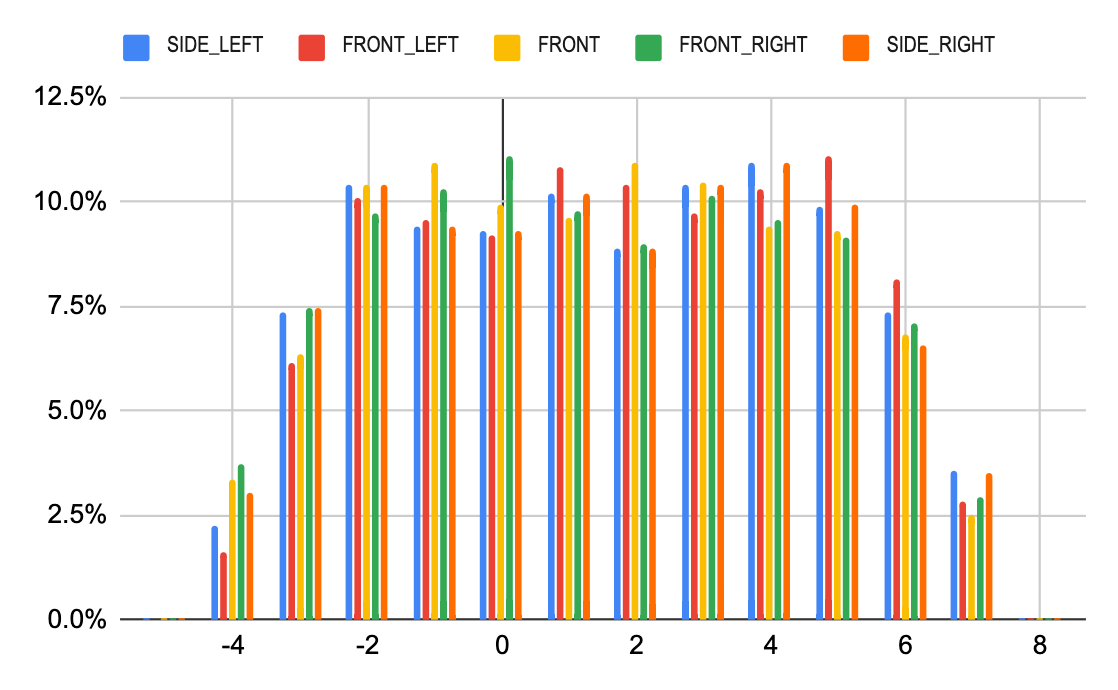}
    \caption{Camera LiDAR synchronization accuracy in milliseconds. The number in x-axis is in milli-seconds. The y-axis denotes the percentage of data frames.\vspace{0cm}}
    \label{fig:sync_accuracy}
\end{figure}

Our cameras and LiDARs data are well-synchronized. The synchronization accuracy is computed as
\begin{align}
\begin{split}
    \textrm{camera\_center\_time} - \textrm{frame\_start\_time} - \\ \textrm{camera\_center\_offset} / 360\degree * 0.1s
\end{split}
\end{align}
The camera\_center\_time is the exposure time of the image's center pixel. The frame\_start\_time is the start time of this data frame. The camera\_center\_offset is the offset of the +x axis of each camera sensor frame w.r.t. the backward direction of the vehicle. The camera\_center\_offset is 90\degree for SIDE\_LEFT camera, $90\degree+45\degree$ for FRONT\_LEFT camera etc.
See Figure \ref{fig:sync_accuracy} for the synchronization accuracy for all the cameras. The synchronization error is bounded in [-6ms, 7ms] with 99.7\% confidence, [-6ms, 8ms] with 99.9995\% confidence.

\textbf{Camera images} are JPEG compressed images. Rolling shutter timing information is provided with each image.

\textbf{Rolling shutter projection.} For any given point~$p$ in the global frame, the rolling shutter camera captures the point at an unknown time~$t$. We can estimate the vehicle pose at~$t$ assuming a constant velocity~$v$ and angular velocity~$\omega$. Using the pose at~$t$, we can project~$p$ to the image and get an image point~$q$, which uniquely defines a pixel capture time~$\Tilde{t}$. We minimize the difference between $t$ and $\Tilde{t}$ by solving a single variable~($t$) convex quadratic optimization. The algorithm is efficient and can be used in real time as it usually converges in 2 or 3 iterations. See Figure \ref{fig:cp_ex} for an example output of the projection algorithm.

\subsection{Dataset Analysis}
\label{dataset_analysis}
\begin{figure*}[t!]
    \centering
    \includegraphics[height=0.30\textwidth]{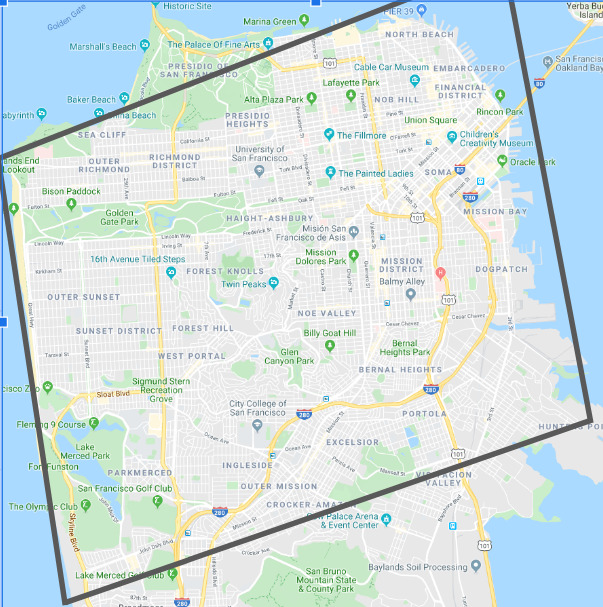}
    \hfill
    \includegraphics[height=0.30\textwidth]{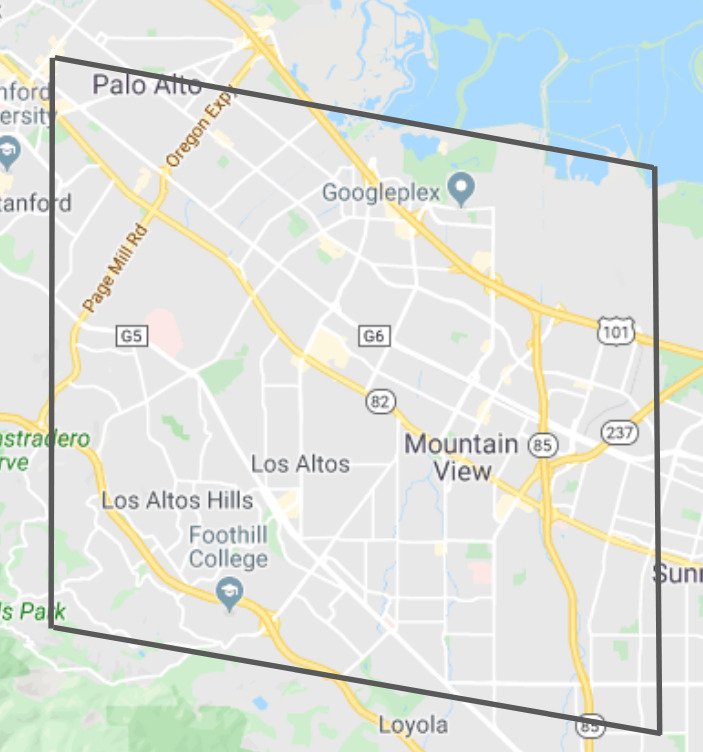}
    \hfill
    \includegraphics[height=0.30\textwidth]{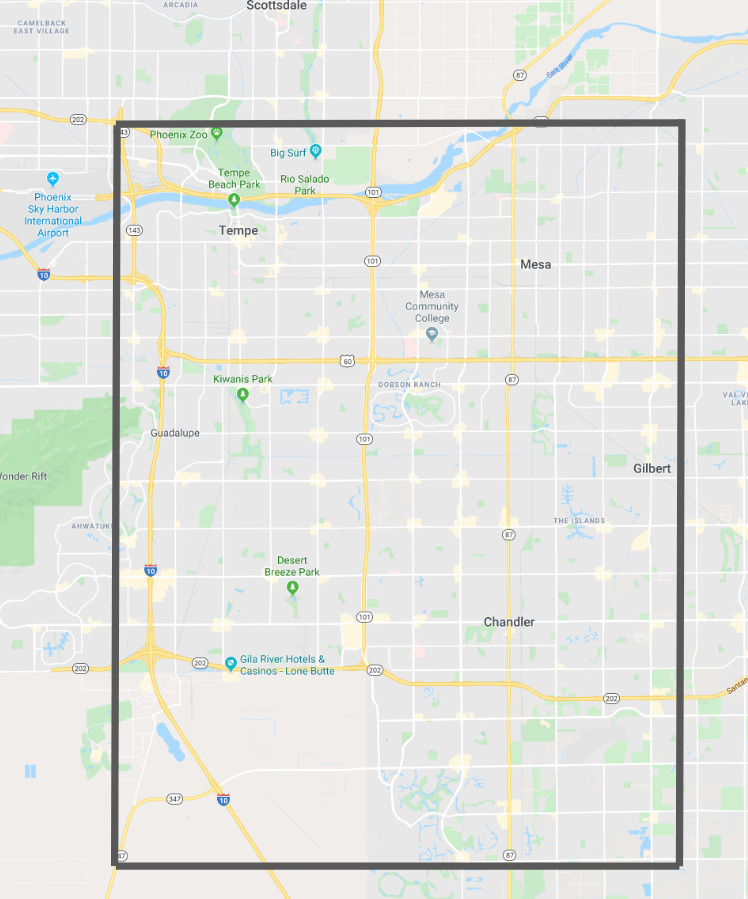}
    \caption{Parallelogram cover of all level 13 S2 cells touched by all ego poses in San Francisco, Mountain View, and Phoenix.\vspace{-0.6cm}}
    \label{fig:map-sf-mtv-phx}
\end{figure*}

The dataset has scenes selected from both suburban and urban areas, from different times of the day. See Table ~\ref{diversity_scene_count}  for the distribution. In addition to the urban/suburban and time of day diversity, scenes in the dataset are selected from many different parts within the cities. We define a geographical coverage metric as the area of the union of all 150-meter diluted ego-poses in the dataset. By this definition, our dataset covers an area of $40\textrm{km}^2$ in Phoenix, and $36\textrm{km}^2$ combined in San Francisco and Mountain View. See Figure~\ref{fig:map-sf-mtv-phx} for the parallelogram cover of all level 13 S2 cells \cite{s2_geometry} touched by all ego poses from all scenes.

\begin{table}[!tbp]
  \centering
  \small{
\begin{tabular}{lccc|ccc}
\toprule
& PHX & MTV & SF & Day & Night & Dawn \\
\midrule
Train & 286 & 103 & 409 & 646 & 79 & 73 \\
Validation & 93 & 21 & 88 & 160 & 23 & 19 \\
\bottomrule
\end{tabular}
}
\caption{Scene counts for Phoenix (PHX), Mountain View (MTV), and San Francisco (SF) and different time of the day for training and validation set. \vspace{-0.4cm}}
\label{diversity_scene_count}
\end{table}


The dataset has around 12M labeled 3D LiDAR objects, around 113k unique LiDAR tracking IDs, around 12M labeled 2D image objects and around 254k unique image tracking IDs. See Table~\ref{num_objects} for counts of each category. 

\begin{table}[!tbp]
  \centering
\small{
\begin{tabular}{lcccc}
\toprule
& Vehicle & Pedestrian & Cyclist & Sign \\ \midrule
3D Object & 6.1M & 2.8M & 67k & 3.2M \\ 
3D TrackID & 60k & 23k & 620 & 23k \\ 
2D Object &  9.0M & 2.7M & 81k & --\\ 
2D TrackID & 194k & 58k &1.7k & --\\
\bottomrule
\end{tabular}
}
\caption{Labeled object and tracking ID counts for different object types. 3D labels are LiDAR labels. 2D labels are camera image labels.\vspace{-0.1cm}}
\label{num_objects}
\end{table}


\section{Tasks}
\label{sec:tasks}

We define 2D and 3D object detection and tracking tasks for the dataset. We anticipate adding other tasks such as segmentation, domain adaptation, behavior prediction, and imitative planning in the future.

For consistent reporting of results, we provide pre-defined training (798 scenes), validation (202 scenes), and test set splits (150 scenes). See Table ~\ref{num_objects} for the number of objects in each labeled category. The LiDAR annotations capture all objects within a radius of 75m. The camera image annotations capture all objects that are visible in the camera images, independent of the LiDAR data. 


\subsection{Object Detection}
\label{object_detection}

\subsubsection{3D Detection}
\label{3d_object_detection}
For a given frame, the 3D detection task involves predicting 3D upright boxes for vehicles, pedestrians, signs, and cyclists. Detection methods may use data from any of the LiDAR and camera sensors; they may also choose to leverage sensor inputs from preceding frames.

Accurate heading prediction is critical for autonomous driving, including tracking and behavior prediction tasks. Average precision (AP), commonly used for object detection, does not have a notion of heading. Our proposed  metric,  APH,  incorporates  heading information into a familiar object detection metric with minimal changes.
\begin{align}
    \textrm{AP} = 100\int_{0}^{1}{\max\{p(r') | r' >= r\} dr}, \\
    \textrm{APH} = 100\int_{0}^{1}{\max\{h(r') | r' >= r\} dr},
\end{align}
where $p(r)$ is the P/R curve. Further, $h(r)$ is computed similar to $p(r)$, but each true positive is weighted by heading accuracy defined as $\min(|\tilde{\theta} - \theta|, 2\pi - |\tilde{\theta} - \theta|)/\pi$, where $\tilde{\theta}$ and ${\theta}$ are the predicted heading and the ground truth heading in radians within $[-\pi, \pi]$. The metrics implementation takes a set of predictions with scores normalized to $[0, 1]$, and samples a fixed number of score thresholds uniformly in this interval. For each score threshold sampled, it does a Hungarian matching between the predictions with score above the threshold and ground truths to maximize the overall IoU between matched pairs. It computes precision and recall based on the matching result. If the gap between recall values of two consecutive operating points on the PR curve is larger than a preset threshold (set to 0.05), more $p/r$ points are explicitly inserted between with conservative precisions. Example: $p(r): p(0) = 1.0, p(1) = 0.0, \delta = 0.05$. We add $p(0.95) = 0.0, p(0.90) = 0.0, ..., p(0.05) = 0.0$. The $\textrm{AP} = 0.05$ after this augmentation. This avoids producing an over-estimated AP with very sparse $p/r$ curve sampling. This implementation can be easily parallelized, which makes it more efficient when evaluating on a large dataset. IoU is used to decide true positives for vehicle, pedestrian and cyclist. Box center distances are used to decide true positives for sign.
\subsubsection{2D Object Detection in Camera Images} In contrast to the 3D~detection task, the 2D camera image detection task restricts the input data to camera images, excluding LiDAR data. The task is to produce 2D axis-aligned bounding boxes in the camera images based on a single camera image. For this task, we consider the AP metric for the object classes of vehicles, pedestrians, and cyclists. We use the same AP metric implementation as described in Section ~\ref{3d_object_detection} except that 2D IoU is used for matching.

\subsection{Object Tracking}
Multi-Object Tracking involves accurately tracking of the identity, location, and optionally properties (e.g. shape or box dimensions) of objects in a scene over time.

Our dataset is organized into sequences, each 20 seconds long with multiple sensors producing data sampled at 10Hz. Additionally, every object in the dataset is annotated with a unique identifier that is consistent across each sequence. We support evaluation of tracking results in both 2D image view, and 3D vehicle centric coordinates.

To evaluate the tracking performance, we use the multiple object tracking (MOT) metric \cite{bernardin2008evaluating}. This metric aims to consolidate several different characteristics of tracking systems -- namely the ability of the tracker to detect, localize, and track the identities of objects over time -- into a single metric to aid in direct comparison of method quality:
\begin{align}
   \textrm{MOTA} &= 100 - 100 \frac{\sum_{t}(m_t + \textrm{fp}_t + \textrm{mme}_t)}{\sum_{t}g_t} \\
    \textrm{MOTP} &= 100\frac{\sum_{i,t}d_t^i}{\sum_{t}c_t}.
\end{align}

Let $m_t$, $\textrm{fp}_t$ and $\textrm{mme}_t$ represent the number of misses, false positives and mismatches. Let $g_t$ be the ground truth count. A mismatch is counted if a ground truth target is matched to a track and the last known assignment was not the track. In MOTP, let $d_t^i$ represent the distance between a detection and its corresponding ground truth match, and $c_t$ be the number of matches found. The distance function used to calculate $d_t^i$ is $1 - \textrm{IoU}$ for a matched pair of boxes. See \cite{bernardin2008evaluating} for the full procedure.

Similar to the detection metrics implementation described in \ref{object_detection}, we sample scores directly and compute an MOTA for each score cutoff. We pick the highest MOTA among all the score cutoffs as the final metric.

\section{Experiments}
\label{sec:expts}
We provide baselines on our datasets based on recent approaches for detection and tracking for vehicles and pedestrians. The same method can be applied to other object types in the dataset. We use 0.7 IoU for vehicles and 0.5 IoU for pedestrians when computing metrics for all tasks.

\subsection{Baselines for Object Detection}
\label{subsec:expts_object_detection}


\paragraph{3D LiDAR Detection}
To establish a 3D Object Detection baseline, we reimplemented PointPillars~\cite{REF:pointpillars_cvpr2018}, which is a simple and efficient LiDAR-based 3D~detector that first uses a single layer PointNet~\cite{REF:Qi2017PointNetDL} to voxelize the point cloud into the Birds Eye View, followed by a CNN region proposal network~\cite{REF:VoxelNet_CVPR2018}. We trained the model on single frame of sensor data with all LiDARs included. 

For vehicles and pedestrians we set the voxel size to 0.33m, the grid range to $[-85\textrm{m}, 85\textrm{m}]$ along the X and Y axes, and $[-3\textrm{m}, 3\textrm{m}]$ along the Z axis. This gives us a $512 \times 512$ pixel Birds Eye View (BEV) pseudo-image. We use the same convolutional backbone architecture as the original paper \cite{REF:pointpillars_cvpr2018}, with the slight exception that our Vehicle model matches our Pedestrian model in having a stride of 1 for the first convolutional block. This decision means both the input and output spatial resolutions of the models are $512 \times 512$ pixels, which increases accuracy at the cost of a more expensive model. We define anchor sizes $(l, w, h)$ as $(4.73\textrm{m}, 2.08\textrm{m}, 1.77\textrm{m})$ for vehicles and $(0.9\textrm{m}, 0.86\textrm{m}, 1.71\textrm{m})$ for pedestrians. Both vehicles and pedestrians have  anchors oriented to $0$ and $\pi/2$ radians. To achieve good heading prediction, we used a different rotation loss formulation, using a smooth-L1 loss of the heading residual error, wrapping the result between $[-\pi, \pi]$ with a huber delta $\delta=\frac{1}{9}$. 

In reference to the LEVEL definition in section ~\ref{ground_truth_labels}, we define the difficulty for the single frame 3D object detection task as follows. We first ignore all 3D labels without any LiDAR points. Next, we assign LEVEL\_2 to examples where either the labeler annotates as hard or if the example has $\leq 5$ LiDAR points. Finally, the rest of the examples are assigned to LEVEL\_1.


We evaluate models on the proposed 3D detection metrics for both 7-degree-of-freedom 3D~boxes and 5-degree-of-freedom BEV boxes on the 150-scene hidden test set.
For our 3D tasks, we use 0.7 IoU for vehicles and 0.5 IoU for pedestrians.
Table~\ref{eval-ap} shows detailed results;

\begin{table*}[h!]
\small{
\begin{center}
\begin{tabular}{l|cccc|cccc}
\toprule
\multirow{2}{*}{Metric} & \multicolumn{4}{c|}{BEV (LEVEL\_1/LEVEL\_2)}              & \multicolumn{4}{c}{3D (LEVEL\_1/LEVEL\_2)}               \\
& Overall & 0 - 30m  & 30 - 50m & 50m - Inf & Overall & 0 - 30m & 30 - 50m & 50m - Inf \\ \midrule
Vehicle APH & 79.1/71.0 & 90.2/87.7 & 77.3/71.1 & 62.8/49.9 & 62.8/55.1 & 81.9/80.8 & 58.5/52.3 & 34.9/26.7 \\
Vehicle AP & 80.1/71.9 & 90.8/88.3 & 78.4/72.2 & 64.8/51.6 & 63.3/55.6 & 82.3/81.2 & 59.2/52.9 & 35.7/27.2 \\ \midrule
Pedestrian APH & 56.1/51.1 & 63.2/61.1 & 54.6/50.5 & 43.9/36.0 & 50.2/45.1 & 59.0/56.7 & 48.3/44.3 & 35.8/28.8 \\
Pedestrian AP & 70.0/63.8 & 76.9/74.5 & 68.5/63.4 & 58.1/47.9 & 62.1/55.9 & 71.3/68.6 & 60.1/55.2 & 47.0/37.9 \\
\bottomrule
\end{tabular}
\end{center}
}
\caption{Baseline APH and AP for vehicles and pedestrians.}\vspace{-0.4cm}
\label{eval-ap}
\end{table*}

\paragraph{2D Object Detection in Camera Images}

We use the Faster R-CNN object detection architecture~\cite{ren2015faster}, with ResNet-101~\cite{he2016deep} as the feature extractor. We pre-trained the model on the COCO Dataset~\cite{lin2014microsoft} before fine-tuning the model on our dataset. We then run the detector on all 5 camera images, and aggregate the results for evaluation. The resulting model achieved an AP of 63.7 at LEVEL\_1 and 53.3 at LEVEL\_2 on vehicles, and an AP of 55.8 at LEVEL\_1 and 52.7 at LEVEL\_2 on pedestrians. 


\subsection{Baselines for Multi-Object Tracking}
\label{subsec:object_tracking}


\paragraph{3D Tracking} We provide an online 3D multi-object tracking baseline following the common tracking-by-detection paradigm, leaning heavily on the above PointPillars \cite{REF:pointpillars_cvpr2018} models. Our method is similar in spirit to \cite{Weng2019_3dmot}. In this paradigm, tracking at each timestep $t$ consists of running a detector to generate detections $d_t^n = \{d_t^1, d_t^2, ..., d_t^n\}$ with $n$ being the total number of detections, associating these detections to our tracks $t_t^m = \{t_t^1, t_t^2, ..., t_t^m\}$ with $m$ being the current number of tracks, and updating the state of these tracks $t_t^m$ given the new information from detects $d_t^n$. Additionally, we need to provide a birth and death process to determine when a given track is Dead (not to be matched with), Pending (not confident enough yet), and Live (being returned from the tracker).

For our baseline, we use our already trained PointPillars \cite{REF:pointpillars_cvpr2018} models from above, $1 - IOU$ as our cost function, the Hungarian method \cite{hungarian_method} as our assignment function, and a Kalman Filter \cite{kalman_filter} as our state update function. We ignore detections with lower than a 0.2 class score, and set a minimum threshold of 0.5 IoU for a track and a detect to be considered a match. Our tracked state consists of a 10 parameter state $t_t^m = \{cx, cy, cz, w, l, h, \alpha, vx, vy, vz\}$ with a constant velocity model. For our birth and death process, we simply increment the score of the track with the associated detection score if seen, decrement by a fixed cost (0.3) if the track is unmatched, and provide a floor and ceiling of the score [0, 3]. Both vehicle and pedestrian results can be seen in Table \ref{eval-bev-mot}. For both vehicles and pedestrians the mismatch percentage is quite low, indicating IoU with a Hungarian algorithm \cite{hungarian_method} is a reasonable assignment method. Most of the loss of MOTA appears to be due to misses that could either be due to localization, recall, or box shape prediction issues.


\begin{table*}[h!]
\small{
\begin{center}
\begin{tabular}{l|ccccc|ccc}
\toprule
\multirow{2}{*}{Metric} & \multicolumn{5}{c|}{Overall (LEVEL\_1/LEVEL\_2)}   & \multicolumn{3}{c}{MOTA by Range (LEVEL\_1/LEVEL\_2)}   \\
            & MOTA & MOTP  & Miss & Mismatch & FP  & 0 - 30m & 30 - 50m & 50m - Inf \\ \midrule


Vehicle 3D & 42.5/40.1 & 18.6/18.6 & 40.0/43.4 & 0.14/0.13 & 17.3/16.4 & 70.6/69.9 & 39.7/37.5 & 12.5/11.2 \\ \midrule


Pedestrian 3D& 38.9/37.7 & 34.0/34.0 & 48.6/50.2 & 0.49/0.47 & 12.0/11.6 & 52.5/51.4 & 37.6/36.5 & 22.3/21.3 \\
\bottomrule
\end{tabular}
\end{center}
}
\caption{Baseline multi-object tracking metrics for vehicles and pedestrians.\vspace{-0.4cm}}
\label{eval-bev-mot}
\end{table*}

\paragraph{2D Tracking} We use the visual multi-object tracking method Tracktor~\cite{Kim18} based on a Faster R-CNN~object detector that we pre-trained on the COCO Dataset~\cite{lin2014microsoft} and then fine-tuned on our dataset. We optimized the parameters of the Tracktor method on our dataset and set $\sigma_\text{active} = 0.4$, $\lambda_\text{active} = 0.6$, and $\lambda_\text{new} = 0.3$. The resulting Tracktor model achieved a MOTA of~34.8 at LEVEL\_1 and~28.3 at LEVEL\_2 when tracking vehicles.
%
%
%

\subsection{Domain Gap}
\label{subsec:domain_adaptation}
The majority of the scenes in our dataset were recorded in three distinct cities (Table ~\ref{diversity_scene_count}), namely San Francisco, Phoenix, Mountain View. We treat Phoenix and Mountain View as one domain called Suburban (SUB) in this experiment. SF and SUB have similar number of scenes per (Table ~\ref{diversity_scene_count}) and different number of objects in total (Table ~\ref{domain_object_count}). As these two domains differ from each other in fascinating ways, the resulting domain gap in our dataset opens up exciting research avenues in the field of domain adaptation. We studied the effects of this domain gap by evaluating the performance of object detectors trained on data recorded in one domain on the training set and evaluated in another domain on the validation set.

We used the object detectors described in Section~\ref{subsec:expts_object_detection}. We filter the training and validation datasets to only contain frames from a specific geographic subset referred to as SF (San Francisco), SUB (MTV + Phoenix), or ALL (all data), and retrain and reevaluate models on the permutation of these splits. Table~\ref{eval-domain-shift} summarizes our results. For the 3D LiDAR-based vehicle object detector, we observed an APH reduction of 8.0 when training on SF and evaluating on SUB compared with training on SUB and evaluating on SUB, and an APH reduction of 7.6 when training on SUB and evaluating on SF compared with training on SF and evaluating on SF. For 3D object detection of pedestrians, the results are interesting. When evaluating on SUB, training on either SF or SUB yield similar APH, while training on all data yields a 7+ APH improvement. This result does not hold when evaluating on SF. Training just on SF when evaluating on SF yields a 2.4 APH improvement as compared to training on the larger combined dataset, while training on SUB only and evaluating on SF leads to a 19.8 APH loss. This interesting behavior on pedestrian might be due to the limited amount pedestrians available in SUB (MTV + Phoenix).
Overall, these results suggest a pronounced domain gap between San Francisco and Phoenix in terms of 3D object detection, which opens up exciting research opportunities to close the gap by utilizing semi-supervised or unsupervised domain adaptation algorithms.

\begin{table}[h!]
\footnotesize
\begin{center}
\begin{tabular}{lcccc}
\toprule
& SF(Tra) & SUB(Tra) & SF(Val) & SUB(Val) \\ \midrule
Vehicle & 2.9M  & 1.9M & 691K & 555K  \\
Pedestrian & 2.0M &  210K & 435K & 103K\\
\bottomrule
\end{tabular}
\end{center}
\caption{3D LiDAR object counts for each domain in training (Tra) and Validation (Val) sets.\vspace{-0.4cm}}
\label{domain_object_count}
\end{table}

\begin{table}[h!]
\footnotesize
\begin{center}
\begin{tabular}{lcc}
\toprule
&ALL/SUB/SF$\rightarrow$SUB &ALL/SF/SUB$\rightarrow$SF \\ \midrule
Vehicle & 45.3/44.0/36.7 & 50.3/49.2/42.5 \\
Pedestrian & 25.7/20.6/19.9 & 46.0/47.6/29.7 \\
\bottomrule
\end{tabular}
\end{center}
\caption{3D object detection baseline LEVEL\_2 APH results for domain shift on 3D vehicles and pedestrians on the \textit{Validation set}. IoU thresholds: Vehicle 0.7, Pedestrian 0.5.\vspace{-0.4cm}}
\label{eval-domain-shift}
\end{table}



\subsection{Dataset Size} 
\label{subsec:dataset_size}
A larger dataset enables research on data intensive algorithms such as Lasernet\cite{meyer2019lasernet}. For methods that work well on small datasets such as PointPillars \cite{REF:pointpillars_cvpr2018}, more data can achieve better results without requiring data augmentation: we trained the same PointPillars model \cite{REF:pointpillars_cvpr2018} from Section \ref{subsec:expts_object_detection} on subsets of the training sequences and evaluated these models on the test set. To have meaningful results, these subsets are cumulative, meaning that the larger subsets of sequences contain the smaller subsets. The results for these experiments can be found in Table \ref{eval-dataset-size}.



\begin{table}[h!]
\footnotesize
\begin{center}
\begin{tabular}{lcccc}
\toprule
Dataset \%-age & 10\% & 30\% & 50\% & 100\%  \\ \midrule
Vehicle & 29.7/28.9 & 41.4/41.0 & 46.3/45.8 & 49.8/49.4 \\
Pedestrian & 39.5/27.7 & 45.7/35.7 & 50.3/40.4 & 53.0/43.0 \\
\bottomrule
\end{tabular}
\end{center}
\caption{The AP/APH at LEVEL\_2 difficulty on the \textit{Validation set} of Vehicles and Pedestrians as the dataset size grows. Each column uses a cumulative random slice of the training set with size determined by the percentage in the first row.\vspace{-0.4cm}}
\label{eval-dataset-size}
\end{table}


\section{Conclusion}
\label{sec:conclusion}

We presented a large-scale multimodal camera-LiDAR dataset that is significantly larger, higher quality, more geographically diverse than any existing similar dataset. It covers $76\textrm{km}^2$  when considering the diluted ego poses at a visibility of 150 meters. We demonstrated domain diversity among Phoenix, Mountain View and San Francisco data in this dataset, which opens exciting research opportunities for domain adaptation. We evaluated the performance of 2D and 3D object detectors and trackers on the dataset. The dataset and the corresponding code are publicly available; we will maintain a public leaderboard to keep track of progress in the tasks. In the future, we plan to add map information, more labeled and unlabeled data with more diversity focused on different driving behaviors and different weather conditions to enable exciting research on other self-driving related tasks, such as behavior prediction, planning and more diverse domain adaptation.


{\small
\bibliographystyle{ieee_fullname}
\bibliography{references}
}

\end{document}